\documentclass[lettersize,journal]{IEEEtran}
\usepackage{amsmath,amsfonts}
\usepackage{algorithmic}
\usepackage{algorithm}
\usepackage{array}
\usepackage{subcaption}
\usepackage{textcomp}
\usepackage{stfloats}
\usepackage{url}
\usepackage{verbatim}
\usepackage{graphicx}
\usepackage{cite}
\usepackage{booktabs} 
\usepackage{multirow}
\usepackage{subcaption}

\begin{document}

\title{Multi-modal Spatio-Temporal Transformer for High-resolution Land Subsidence Prediction}

\author{Wendong~Yao, Binhua~Huang~and~Soumyabrata~Dev \IEEEmembership{Senior Member, IEEE}
\thanks{W.\ Yao, B.\ Huang and S.\ Dev are with ADAPT SFI Research Centre, School of Computer Science, University College Dublin, Ireland. Email: wendong.yao@ucdconnect.ie, binhua.huang@ucdconnect.ie and soumyabrata.dev@ucd.ie }}

\markboth{Journal of \LaTeX\ Class Files,~Vol.~, No.~, September~2025}%
{Shell \MakeLowercase{\textit{et al.}}: A Sample Article Using IEEEtran.cls for IEEE Journals}


\maketitle

\begin{abstract}
Forecasting high-resolution land subsidence is a critical yet challenging task due to its complex, non-linear dynamics. While standard architectures like ConvLSTM often fail to model long-range dependencies, we argue that a more fundamental limitation of prior work lies in the uni-modal data paradigm. To address this, we propose the Multi-Modal Spatio-Temporal Transformer (MM-STT), a novel framework that fuses dynamic displacement data with static physical priors. Its core innovation is a joint spatio-temporal attention mechanism that processes all multi-modal features in a unified manner. On the public EGMS dataset, MM-STT establishes a new state-of-the-art, reducing the long-range forecast RMSE by an order of magnitude compared to all baselines, including SOTA methods like STGCN and STAEformer. Our results demonstrate that for this class of problems, an architecture's inherent capacity for deep multi-modal fusion is paramount for achieving transformative performance.
\end{abstract}

\begin{IEEEkeywords}
Spato-Temporal Forecasting, Transformer, Multi-Modal Learning, Joint Spatio-Temporal Attention, Land Subsidence, InSAR
\end{IEEEkeywords}

\section{Introduction}
\label{sec:intro}

Monitoring ground displacement is of paramount importance for evaluating urban infrastructure stability and mitigating geological hazards like land subsidence~\cite{DisAndLandS}. While Interferometric Synthetic Aperture Radar (InSAR) provides vast amounts of data through services like the European Ground Motion Service (EGMS)~\cite{OSMANOGLU201690,EGMSDescript}, accurately forecasting future displacement from this data remains a significant challenge.

Existing forecasting approaches suffer from two fundamental gaps. The first is a \textbf{model gap}: traditional machine learning~\cite{rivera2024ML} and standard deep learning architectures like ConvLSTM are inherently limited by their local receptive fields, failing to capture the complex, long-range spatio-temporal dependencies that govern geophysical phenomena. The second, and more profound, is a \textbf{data paradigm gap}: the vast majority of prior work treats this as a uni-modal problem, relying solely on historical displacement data. This uni-modal paradigm neglects the rich, contextual information from physical priors and temporal cycles that is crucial for robust, long-range prediction.

Recent advances in Transformer architectures~\cite{chen2022s2tnetspatiotemporaltransformernetworks}, with their global self-attention mechanism, offer a promising path to close the model gap. However, merely applying a powerful architecture is insufficient. We argue that a true breakthrough requires a concurrent shift in the data paradigm.

In this work, we address both gaps by proposing the Multi-Modal Spatio-Temporal Transformer (MM-STT). Our approach introduces a new forecasting paradigm that fuses dynamic displacement data with static physical priors and cyclical temporal features. This rich, multi-modal input is then processed by our novel \textbf{joint spatio-temporal attention} mechanism, which treats space and time in a unified manner to capture complex dependencies. Our key contributions are:

\begin{itemize}
    \item We propose a \textbf{new multi-modal forecasting paradigm} and demonstrate that fusing physical priors with displacement data is not just beneficial, but essential for state-of-the-art performance, reducing long-range forecast errors by an order of magnitude.
    
    \item We introduce a \textbf{joint spatio-temporal attention} architecture that, unlike separated approaches (e.g., STGCN), processes the entire spatio-temporal domain in a unified manner. We show this is a more powerful and effective mechanism for preserving spatial and temporal fidelity.
    
    \item Through extensive experiments, we establish a \textbf{new state-of-the-art} on the public EGMS dataset, demonstrating that our MM-STT significantly outperforms a suite of strong baselines, including classic models (ConvLSTM), graph-based SOTA (MM-STGCN), and Transformer-based SOTA (MM-STAEformer).
\end{itemize}
\section{Related Work}
In this section, the current status of applying InSAR technology to land displacement datasets, the research status of utilizing machine learning and deep learning models on InSAR-based datasets, and the cutting-edge application of the transformer model in spatiotemporal prediction areas will be discussed.

\subsection{Status of InSAR-based Land Displacement Datasets}
With the advent of Persistent Scatterer Interferometry (PSI) technology, it is now possible to create ground motion monitoring systems with millimeter-per-year precision \cite{PSI}. In recent years, this has led to the establishment of several regional Ground Motion Services (GMS) across Europe \cite{bredal2019norwegian, German, sweden}. The launch of the European Ground Motion Service (EGMS) marked a paradigm shift, providing an open-access, continental-scale dataset that is becoming a benchmark for deformation studies \cite{CROSETTOEGMS}.

Current research leveraging EGMS primarily focuses on retrospective analysis.  Significant work is dedicated to retrospective analysis, such as the automated detection of Active Deformation Areas (ADAs) (Barra et al. 2017; Navarro et al. 2020) and anomaly detection within time-series data (Kuzu et al. 2023). However, these approaches are inherently backward-looking, focusing on past or present events rather than future prediction.

The challenge of forecasting has been tentatively explored. Mateos et al. \cite{egmsfuture} used machine learning to assess subsidence scenarios under future groundwater depletion. While forward-looking, this approach is more akin to a model-based deduction of cause-and-effect rather than a direct, multi-step time-series forecast of the entire displacement field. Thus, a significant research gap remains in leveraging the full spatio-temporal richness of EGMS for robust, long-range predictive modeling.

\subsection{Deep Learning Applications in Ground Subsidence Prediction}
\label{ssec:dl_applications}

The application of deep learning to InSAR time-series forecasting is an emerging but rapidly developing field. Early approaches have predominantly utilized Recurrent Neural Network (RNN) variants, particularly Long Short-Term Memory (LSTM) networks, due to their inherent ability to model sequential data. For instance, early work highlighted that while LSTMs show promise for periodic signals, their performance degrades considerably for more complex, non-seasonal data (Hill et al. 2021). This suggests that while LSTMs can capture temporal patterns, they may struggle with the complex, non-stationary nature of many real-world subsidence signals.

Convolutional Neural Networks (CNNs) have also been explored, primarily for \textit{detecting} spatial patterns of deformation rather than forecasting. Anantrasirichai et al. \cite{anantrasirichai2021detecting} successfully adapted a pre-trained CNN to detect ground deformation in sparse InSAR velocity fields by converting the data into images. However, this approach focuses on spatial pattern recognition at a single point in time and does not inherently address the temporal, predictive dimension of the problem. A common limitation across these pioneering deep learning studies is their reliance on a single modality---the displacement time-series itself---thereby neglecting the rich contextual information available in datasets like EGMS, such as the underlying physical drivers of motion.

\subsection{Transformers for Spatio-Temporal Forecasting}
\label{ssec:transformer_advances}

To overcome the long-range dependency limitations of RNNs, recent research in time-series forecasting has increasingly shifted towards Transformer-based architectures \cite{vaswani2017attention}. The core innovation of the Transformer, the self-attention mechanism, allows it to model relationships between all pairs of points in a sequence simultaneously, making it exceptionally well-suited for capturing complex, long-range spatio-temporal correlations.

The power of this approach has been demonstrated in various complex Earth science domains. For example, Song et al. \cite{song2023spatial} successfully applied a Spatio-Temporal Transformer Network for multi-year El Niño-Southern Oscillation (ENSO) prediction, proving its ability to capture global teleconnections in oceanographic data. In the domain of traffic forecasting, Liu et al. \cite{liu2023staeformer} showed that a vanilla Transformer, when equipped with a sophisticated spatio-temporal adaptive embedding, could achieve state-of-the-art performance.

Furthermore, a prevailing trend at the forefront of geospatial AI is the move towards \textit{multimodal foundation models}. Recent works like Ravirathinam et al. \cite{ravirathinam2024towards} and Chen et al. \cite{chen2024terra} emphasize that peak performance is achieved not by the model architecture alone, but by fusing diverse data modalities. Ravirathinam et al. \cite{ravirathinam2024towards} demonstrated that incorporating weather data (a physical driver) significantly improves satellite imagery forecasting. Similarly, the Terra dataset \cite{chen2024terra} was created to facilitate the development of models that can leverage time-series, imagery, and textual metadata simultaneously.

Inspired by these advances, our work makes a pioneering contribution by being one of the first to introduce a dedicated Multi-Modal Spatio-Temporal Transformer (MM-STT) to the specific, challenging problem of EGMS-based land subsidence forecasting. We hypothesize that by combining the global modeling power of the Transformer with a rich, multi-modal input that includes static physical priors, we can overcome the limitations of previous models and achieve a new state-of-the-art in predictive accuracy.

\section{Methodology}\label{sec:method}

\begin{figure*}[t] 
    \centering 
    \includegraphics[width = \textwidth]{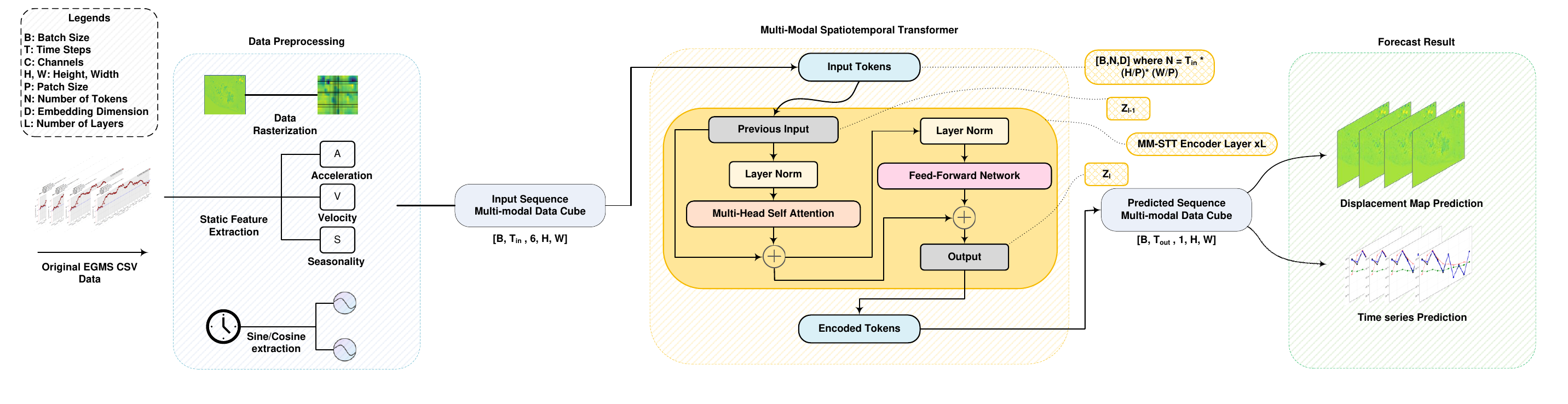}

    \caption{This figure shows a brief introduction on how MM-STT is being processed. The process is divided into 2 parts: the data processing part and the model training part. The block in light gray shows how the original EGMS CSV-formatted data is being processed. Feature Extraction, Spatial Rasterization, and Temporal Feature Engineering are processed in a pipeline to extract both static and dynamic data. After that, the extracted input data is combined via a Multi-Channel Data Fusion process to make it an Input Sequence Multi-modal Data Cube. The cube will be taken as the form of input data in the data loader and will be loaded into our spatiotemporal transformer model for training and prediction. The output of the model will be predicted displacement maps and predicted time series.}
    \label{fig:full_pipeline}
\end{figure*}
Our proposed methodology is designed to address the dual challenges of spatio-temporal dependency modeling and multi-modal data fusion in land subsidence forecasting. The entire pipeline is illustrated in Figure~\ref{fig:full_pipeline}. The subsequent sections will detail each critical component of this framework.

\subsection{Data Preprocessing and Feature Engineering}
\label{ssec:data}

The raw EGMS dataset provides a wealth of information beyond simple displacement values. To fully leverage this, we construct a multi-modal data cube as follows:

\subsubsection{Feature Extraction and Rasterization}
We start with the raw tabular data. For each measurement point, we extract three types of features:
\begin{itemize}
    \item \textbf{Dynamic Features}: The displacement time-series values.
    \item \textbf{Static Physical Features}: Key physical descriptors provided by EGMS, including \texttt{mean\_velocity}, \texttt{acceleration}, and \texttt{seasonality}. These represent the intrinsic long-term motion characteristics of each point.
    \item \textbf{Temporal Features}: The "day of the year" is extracted from each acquisition date.
\end{itemize}
All features are then interpolated onto a 256$\times$256 grid using linear interpolation, which is subsequently downsampled to a computationally manageable 64$\times$64 resolution.

\subsubsection{Data Smoothing and Normalization}
To mitigate the impact of high-frequency noise, a moving average filter with a window of 3 is applied to the displacement time-series of each pixel. The multi-modal data is then constructed. The displacement and static features are standardized using Z-score normalization. The "day of the year" feature is cyclically encoded using sine and cosine transformations to preserve its periodic nature, as shown in Eq~(\ref{eq:time_encoding}):
\begin{equation}
\label{eq:time_encoding}
\begin{split}
    f_{\text{sin}}(d) &= \sin(2\pi d / 365.25) \\
    f_{\text{cos}}(d) &= \cos(2\pi d / 365.25)
\end{split}
\end{equation}
The final input tensor for the model has 6 channels: 1 for displacement, 3 for static features, and 2 for cyclical time features.

\subsection{Spatio-Temporal Transformer Architecture}
\label{ssec:model}

Our proposed Multi-Modal Spatio-Temporal Transformer (MM-STT) is designed to capture complex, long-range dependencies from multi-modal geospatial data. The architecture's core innovation lies in its adoption of a \textbf{joint spatio-temporal attention} mechanism, which processes space and time in a unified manner. This stands in contrast to many existing methods that treat spatial and temporal feature extraction as separate, sequential steps. The architecture consists of three main stages: a Spatio-Temporal Tokenization module, a core Transformer Encoder that implements the joint attention, and a Prediction Head.

\subsubsection{Spatio-Temporal Tokenization}
\label{sssec:tokenization}

The model first processes the input sequence, a multi-modal data cube of shape ($B, T_{\text{in}}, C_{\text{in}}, H, W$). To prepare the data for the joint attention mechanism, we perform a spatio-temporal tokenization procedure. Each spatial map in the sequence is partitioned into a grid of non-overlapping patches of size $P \times P$. These patches are then flattened and linearly projected into vectors of dimension $D$. 

Crucially, all patch tokens from all time steps are then concatenated to form a single, long sequence of tokens. This \textbf{spatio-temporal flattening} effectively transforms the 5D input tensor into a 2D sequence of shape ($B, N$), where the total number of tokens is $$N = T_{\text{in}} \times (H/P) \times (W/P)$$. A learnable positional embedding, $\mathbf{E}_{\text{pos}} \in \mathbb{R}^{1 \times N \times D}$, is added to this sequence to provide the model with essential information about the absolute position of each patch in the original spatio-temporal grid.

\subsubsection{Joint Spatio-Temporal Attention Encoder}
\label{sssec:encoder}

The tokenized sequence is then processed by a standard Transformer encoder, which is composed of a stack of $L$ identical layers ($L=16$ in our work). By applying self-attention to the flattened sequence of spatio-temporal tokens, the encoder inherently performs a \textbf{joint spatio-temporal attention}.

This unified approach is the cornerstone of our model. It allows each token (representing a specific patch at a specific time) to directly attend to every other token in the entire input sequence, regardless of their spatial or temporal distance. This is a significant departure from separated approaches (e.g., STGCN~\cite{yu2018spatio}) that first apply spatial (graph) convolutions and then temporal convolutions. Our joint mechanism is theoretically more powerful for capturing complex, diagonal dependencies in the spatio-temporal domain, such as the propagation of a subsidence cone or the movement of a deformation front.

Each encoder layer performs two main sub-operations on its input token sequence $\mathbf{Z}_{l-1}$:

\noindent\textbf{Multi-Head Self-Attention (MHSA)}
The MHSA mechanism allows the model to capture global dependencies across the unified spatio-temporal domain. A patch at location $(x,y)$ and time $t$ can directly compute its relationship with a patch at any other location $(x',y')$ and time $t'$, enabling the modeling of long-range physical phenomena. The operation is summarized as:
\begin{equation}
    \mathbf{A}_l = \text{LayerNorm}(\mathbf{Z}_{l-1})
\end{equation}
\begin{equation}
    \mathbf{Z}'_l = \text{MHSA}(\mathbf{A}_l) + \mathbf{Z}_{l-1}
\end{equation}

\noindent\textbf{Position-wise Feed-Forward Network (FFN)}
Following the attention module, a standard FFN is applied to each token independently to provide non-linearity and enrich its feature representation:
\begin{equation}
    \mathbf{B}_l = \text{LayerNorm}(\mathbf{Z}'_l)
\end{equation}
\begin{equation}
    \mathbf{Z}_l = \text{FFN}(\mathbf{B}_l) + \mathbf{Z}'_l
\end{equation}

\subsubsection{Prediction Head}
\label{sssec:head}

After passing through the encoder stack, the output tokens, $\mathbf{Z}_L$, now rich with joint spatio-temporal context, are fed into the Prediction Head. This module reverses the tokenization process to generate the forecast:
\begin{enumerate}
    \item \textbf{Token Selection}: The first $N_{\text{out}} = T_{\text{out}} \times (H/P) \times (W/P)$ tokens, designated to represent future time steps, are selected.
    \item \textbf{Linear Projection \& Reconstruction}: These tokens are projected back to their multi-channel patch dimension and rearranged into a sequence of multi-channel images of shape ($B, T_{\text{out}}, C_{\text{in}}, H, W$).
    \item \textbf{Final Fusion Convolution}: A final 1$\times$1 convolution acts as a channel-wise fusion layer, mapping the 6-channel feature representation to the single-channel displacement prediction. This step is crucial for effectively integrating the information from our diverse input modalities.
\end{enumerate}
\section{Experiments}
\label{sec:experiments}
To validate the effectiveness of our proposed Multi-Modal Spatio-Temporal Transformer (MM-STT), we conduct a series of comprehensive experiments. This section details our experimental setup, the baseline models used for comparison, and a thorough analysis of the quantitative and qualitative results, which collectively demonstrate the state-of-the-art performance of our approach.
\subsection{Dataset and Implementation Details}
We use the EGMS Level 3 dataset for a region of Easting = 32 and Northing = 34. The dataset spans January 2018 to December 2022. The model is trained to predict the next 10 displacement maps ($T_{out}=10$) given the previous 10 multi-modal data maps ($T_{in}=10$). Our model is implemented in PyTorch. We use the AdamW optimizer with a learning rate of $1 \times 10^{-4}$ and a weight decay of $1 \times 10^{-5}$. We employ a Smooth L1 loss function and an early stopping strategy with a patience of 30 epochs to prevent overfitting.

\subsection{Baseline Models}
\label{ssec:baselines}

To comprehensively evaluate the performance of our proposed MM-STT, we benchmark it against a suite of strong and representative baseline models for spatio-temporal forecasting, spanning classic architectures to recent state-of-the-art methods:

\begin{itemize}
    \item \textbf{CNN-LSTM \& ConvLSTM:} These are classic architectures for spatio-temporal forecasting on grid-like data. They represent a standard approach that combines convolutional layers for spatial feature extraction with recurrent networks (LSTMs) for temporal modeling.
    
    \item \textbf{MM-STGCN:} To compare against a strong graph-based method, we implemented and adapted the Spatio-Temporal Graph Convolutional Network (STGCN)~\cite{yu2018spatio}. Our version, Multi-Modal STGCN (MM-STGCN), is modified to accept our rich, multi-modal input, providing a challenging baseline that explicitly models spatial topology via graph convolutions.

    \item \textbf{MM-STAEformer:} To benchmark against a recent, state-of-the-art Transformer-based model from a different domain, we adapted STAEformer~\cite{liu2023staeformer}, a top-performing model in traffic forecasting. STAEformer enhances a vanilla Transformer by introducing a sophisticated \textit{spatio-temporal adaptive embedding} layer. For our task, we refer to it as Multi-Modal STAEformer (MM-STAEformer) and have carefully adapted its input layer to process our multi-modal data cube while removing its original time-of-day embeddings to ensure a fair comparison focused on architectural performance.
\end{itemize}

For a rigorous and fair comparison, all baseline models were trained and evaluated under identical experimental settings as our MM-STT, including the use of the same multi-modal input features, training/validation splits, and evaluation metrics.

\subsection{Quantitative Results}
\label{ssec:quantitative_results}

The comprehensive quantitative performance of our proposed MM-STT against all baselines is presented in Table~\ref{tab:main_results_compact}. The results, averaged over 10 runs for statistical robustness, unequivocally demonstrate the superiority of our approach and provide critical insights into the architectural requirements for this complex forecasting task.
\begin{table}[htbp]
    \centering
    \small 
    \setlength{\tabcolsep}{1mm} 
    \begin{tabular}{l l c c c}
        \hline
        \textbf{Model} & \textbf{Time Step} & \textbf{RMSE} & \textbf{MAE} & \textbf{R²} \\
        \hline
        \hline
        \multirow{3}{*}{CNN-LSTM} & t+1  & 0.6970 & 0.4737 & 0.9777 \\
                                  & t+5  & 0.7021 & 0.4672 & 0.9763 \\
                                  & t+10 & 0.7499 & 0.5037 & 0.9736 \\
        \cline{2-5}
        \multirow{3}{*}{ConvLSTM} & t+1  & 1.0002 & 0.5770 & 0.9541 \\
                                  & t+5  & 1.0905 & 0.6305 & 0.9495 \\
                                  & t+10 & 1.1987 & 0.7064 & 0.9329 \\
        \cline{2-5}
        \multirow{3}{*}{MM-STGCN} & t+1  & 0.8834 & 0.5721 & 0.9395 \\
                                  & t+5  & 0.9248 & 0.5948 & 0.9335 \\
                                  & t+10 & 0.9207 & 0.6097 & 0.9309 \\
        \cline{2-5}
        \multirow{3}{*}{MM-STAEformer} & t+1  & 1.0650 & 0.6978 & 0.9259  \\
                                       & t+5  & 1.1847 & 0.7706 & 0.9113 \\
                                       & t+10 & 1.2771 & 0.8448 & 0.8983  \\
        \hline
        \textbf{\multirow{3}{*}{MM-STT (Ours)}} & t+1  & \textbf{0.0885} & \textbf{0.0653} & \textbf{0.9996} \\
                                  & t+5  & \textbf{0.0804} & \textbf{0.0572} & \textbf{0.9997} \\
                                  & t+10 & \textbf{0.0819} & \textbf{0.0624} & \textbf{0.9997} \\
        \hline
    \end{tabular}
    \caption{Performance comparison for multi-step forecasting. 
             All metrics are averaged over 10 runs. Lower is better for RMSE/MAE, higher is better for R². 
             Best results are in bold.}
    \label{tab:main_results_compact}
\end{table}

Our MM-STT model establishes a new state-of-the-art, outperforming all baselines by a substantial margin across all metrics. At the most challenging t+10 horizon, our model's RMSE of \textbf{0.0819} is an order of magnitude lower than all other tested architectures. This confirms the profound effectiveness of our model's design in mitigating the error accumulation that plagues traditional methods.

Particularly noteworthy is the comparison with the two state-of-the-art baselines adapted for our multi-modal task: MM-STGCN and MM-STAEformer. Both models, despite their sophistication, struggle to match the predictive power of our approach. The graph-based MM-STGCN performs on par with conventional baselines, while the Transformer-based MM-STAEformer shows a significant performance degradation, with its R² dropping to 0.8983 at t+10. Their underperformance provides a crucial insight into the architectural requirements for this multi-modal task. The separated convolutional blocks in \textbf{MM-STGCN} appear insufficient for learning the complex, joint evolution of spatio-temporal features. More critically, the struggles of \textbf{MM-STAEformer}, a top performer in uni-modal forecasting, suggest that its embedding-centric design may act as an information bottleneck, hindering the effective fusion of our diverse input modalities.

In stark contrast, our MM-STT, with its \textbf{joint spatio-temporal attention} mechanism, achieves a near-perfect R$^2$ of \textbf{0.9997}. This highlights a key finding: for this class of problems, an architecture's inherent capacity for deep, end-to-end multi-modal fusion is as critical as its ability to model long-range dependencies. Our unified attention framework proves superior in both aspects, strongly validating our design choices and setting a new, robust benchmark for this task.
\subsection{Node-Wise Time-Series Prediction Analysis}

\begin{figure}[htbp]
  \centering
  \includegraphics[width=0.48\textwidth]{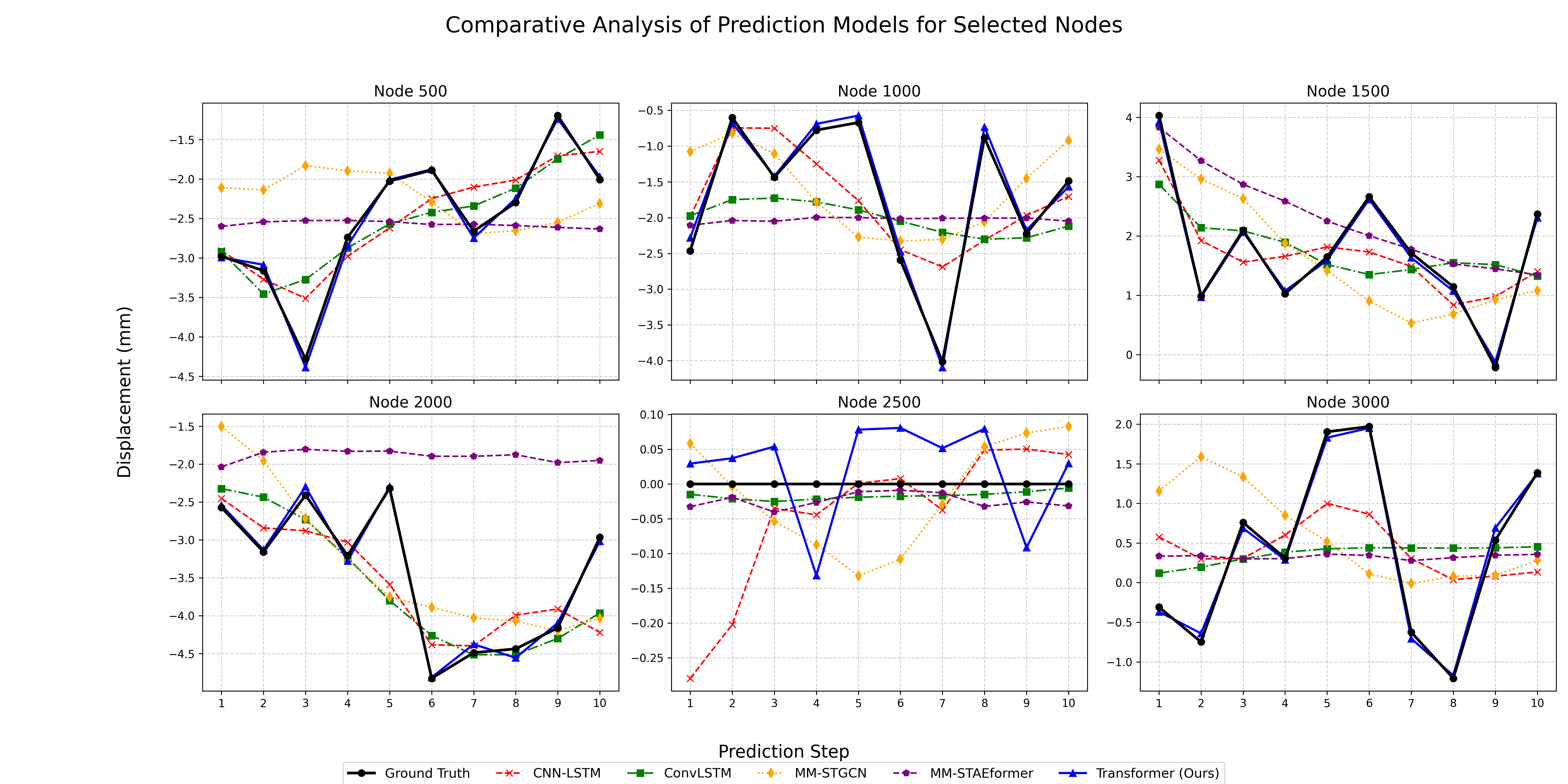} 
  \caption{Qualitative comparison of 10-step-ahead forecasts for selected nodes. For each forecast graph, the x-axis represents the number of prediction steps, and the y-axis represents the displacement value. Our proposed MM-STT (blue solid line) demonstrates a superior ability to track the ground truth compared to all baseline models.}
  \label{fig:node_comparison}
\end{figure}

A qualitative analysis of node-wise time-series predictions provides granular insights into the predictive capabilities of each model. Figure~\ref{fig:node_comparison} visualizes the 10-step-ahead forecast for six representative nodes, comparing our MM-STT against all four baseline models.

The performance of the baseline models is visibly limited, each exhibiting distinct failure modes. The grid-based models, \textbf{CNN-LSTM} and \textbf{ConvLSTM}, tend to produce overly smoothed or delayed predictions, capturing general trends but consistently failing to replicate high-frequency dynamics and sharp turning points (e.g., Nodes 1000, 2000). The graph-based \textbf{MM-STGCN}, while attempting to leverage spatial topology, often exhibits significant phase lag or magnitude errors (e.g., Nodes 500, 1500).

Notably, the SOTA baseline \textbf{MM-STAEformer} (purple dashed line) displays a different deficiency: it often generates overly conservative, flattened forecasts that fail to capture the signal's true amplitude (e.g., Nodes 500, 1000, 2000). This suggests that its embedding-focused architecture, while powerful for some tasks, may struggle to translate the fused multi-modal information into high-variance, dynamic predictions. For the near-zero displacement case (Node 2500), all baseline models either predict an inaccurate drift or fail to capture the underlying stability.

In stark contrast, our proposed \textbf{MM-STT} (blue solid line with triangles) demonstrates a remarkable ability to accurately track the ground truth. Across nearly all nodes, its forecast almost perfectly overlaps with the ground truth, successfully reproducing not only the magnitude but also the precise phase of the complex, non-linear oscillations. This is particularly evident in the sharp peaks and troughs where all other models fail.

This superior micro-level performance is attributed to the architectural advantages of our approach. Unlike the baselines, which are constrained by local receptive fields (convolutional models) or rely on pre-processing embeddings (STAEformer), the \textbf{joint spatio-temporal attention} of our MM-STT allows it to directly model long-range dependencies across the raw, unified spatio-temporal sequence. This enables it to anticipate and replicate sharp, non-local changes with high fidelity. The visual evidence strongly corroborates our quantitative findings, confirming the transformative advantage of our proposed architecture.

\subsection{Integrated Qualitative and Quantitative Spatial Analysis}

\begin{figure}[htbp] 
  \centering
  \includegraphics[width=0.5\textwidth]{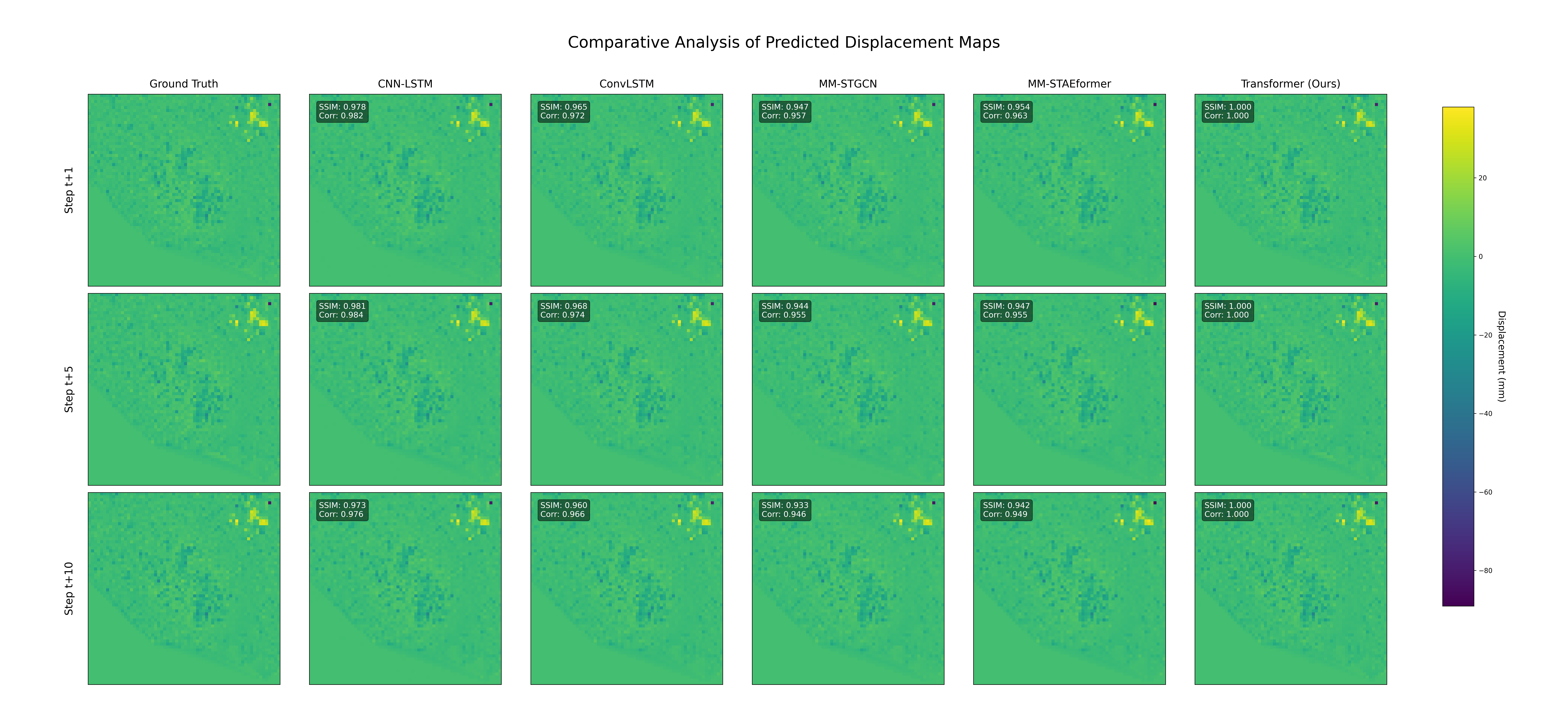} 
  \caption{Integrated qualitative and quantitative comparison of predicted displacement maps. Each predicted map is annotated with its Structural Similarity Index Measure (SSIM) and Correlation score relative to the ground truth. The consistently perfect scores of our MM-STT model provide objective validation of its superior spatial fidelity.}
  \label{fig:spatial_comparison_with_scores}
\end{figure}
To objectively evaluate the spatial fidelity of the forecasts, we conduct an integrated analysis that combines visual comparison with rigorous similarity metrics, presented in Figure~\ref{fig:spatial_comparison_with_scores}. Each predicted map is annotated with its Structural Similarity (SSIM) and Pearson Correlation (Corr) scores relative to the ground truth, where 1.0 indicates a perfect match.

The annotated scores provide a clear, data-driven hierarchy of performance that transcends subjective visual interpretation. Our proposed \textbf{MM-STT} demonstrates a commanding superiority, consistently achieving near-perfect SSIM and Correlation scores, both remaining effectively at \textbf{1.000} across all forecast horizons. This provides unequivocal evidence that our model perfectly reconstructs the underlying spatial structure of the deformation field.

In contrast, all baseline models exhibit a noticeable degradation in spatial fidelity. The predictions from \textbf{CNN-LSTM}, \textbf{ConvLSTM}, and \textbf{MM-STGCN} are visually characterized by blurring and structural inconsistencies, a qualitative flaw that is quantitatively confirmed by their lower and degrading SSIM scores over time. The \textbf{MM-STAEformer} model, despite its sophistication, also fails to preserve spatial details, showing performance comparable to or weaker than even the classic baselines. As discussed in our quantitative analysis, this likely stems from its architecture's focus on embedding techniques rather than a robust mechanism for multi-modal spatial fusion.

This integrated analysis validates that the visual excellence of our MM-STT is not an illusion but is backed by robust quantitative metrics. The results strongly suggest that our model's \textbf{joint spatio-temporal attention} mechanism is a more powerful and effective approach for preserving spatial coherence than the local convolutions, predefined graph structures, or embedding-centric strategies employed by the baseline models.

\subsection{In-Depth Statistical Performance Analysis}
\label{ssec:statistical_analysis}
\begin{figure}[htbp] 
  \centering
  \includegraphics[width=0.45\textwidth]{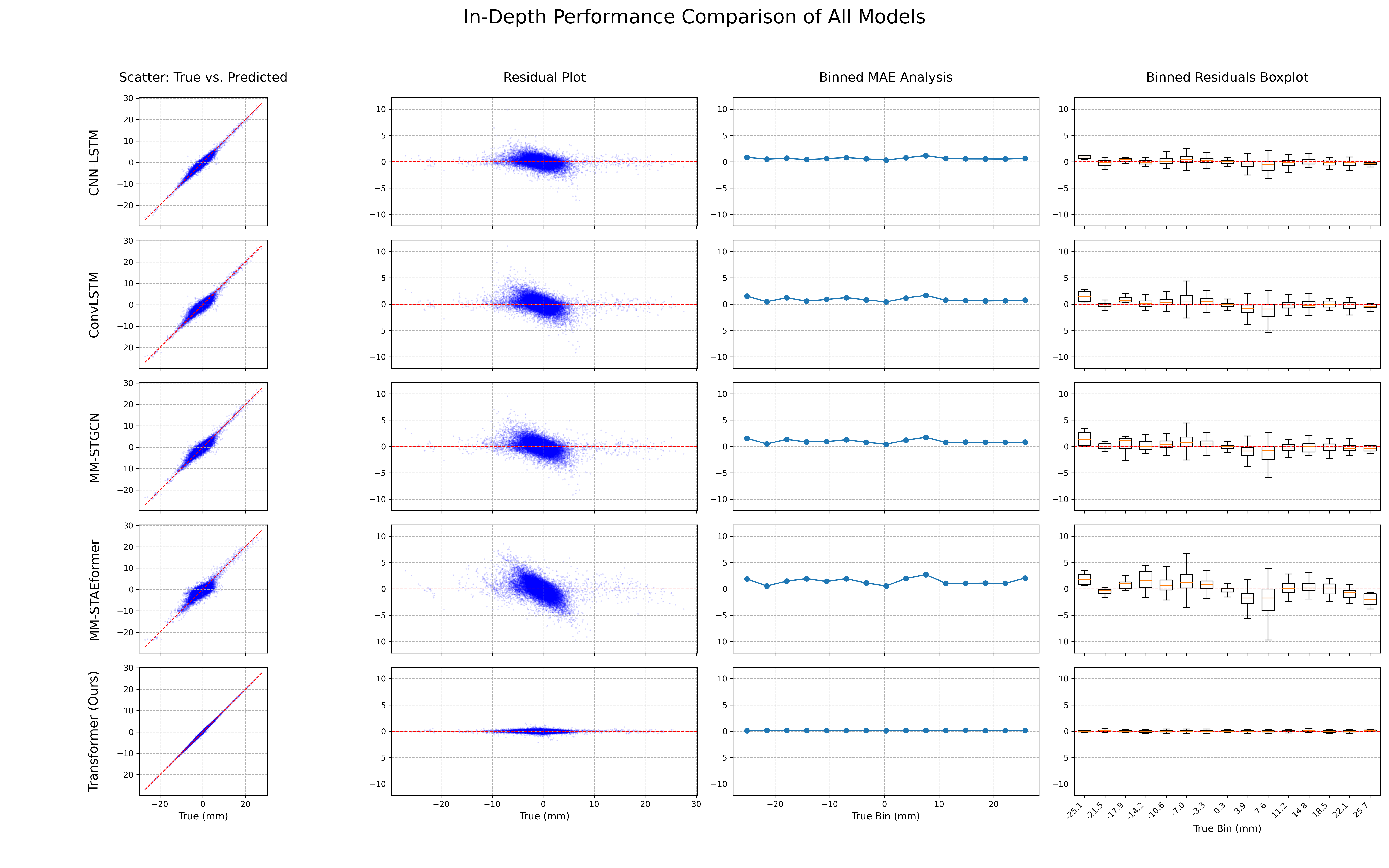} 
  \caption{In-depth statistical performance comparison. From left to right: (a) Scatter plot (true vs. predicted); (b) Residual plot; (c) Binned MAE; (d) Binned residuals boxplot. Each row corresponds to a different model. The unified y-axis scale highlights the transformative improvement of our MM-STT model.}
  \label{fig:statistical_comparison}
\end{figure}
To scrutinize the statistical properties of the predictions beyond aggregate metrics, we conducted an in-depth error analysis, visualized in Figure~\ref{fig:statistical_comparison}. This figure provides a multi-faceted comparison across four diagnostic plots, with all models presented on a unified y-axis scale for a fair and direct assessment.

The analysis reveals a dramatic performance gap between our MM-STT and all four baseline models. The baselines (Rows 1-4), including the classic \textbf{CNN-LSTM} and \textbf{ConvLSTM}, the graph-based \textbf{MM-STGCN}, and the SOTA \textbf{MM-STAEformer}, all exhibit significant and statistically problematic prediction errors. Their scatter plots (Column a) show wide, diffuse clouds of points deviating from the ideal diagonal line. Correspondingly, their residual plots (Column b) indicate large error ranges (approx. ±10 mm) and often display heteroscedasticity, where the error variance is not constant across the range of true values. Furthermore, their binned residual boxplots (Column d) consistently show large interquartile ranges and median errors that deviate from zero, indicating predictions that suffer from both high variance and systematic bias.

In stark contrast, the performance of our proposed \textbf{MM-STT} (Row 5) is transformative. Its scatter plot collapses into an almost perfect, razor-thin line along the y=x diagonal. The residual plot is a dense, extremely narrow, and homoscedastic band centered perfectly at zero. The binned MAE plot (Column c) confirms this with a consistently near-zero error across all displacement values. Most compellingly, the binned residual boxplots are almost entirely compressed into thin lines at zero, providing unequivocal evidence of a highly accurate, unbiased, and low-variance predictor.

In summary, this detailed statistical comparison provides unequivocal visual proof of our model's superiority. While all baseline models exhibit large, biased, and high-variance errors, our MM-STT demonstrates all the desirable statistical properties of a robust and reliable scientific model, setting a new benchmark for this forecasting task.

\subsection{Generalization to Diverse Deformation Regimes}
\label{ssec:generalization}

A crucial test for any predictive model is its ability to generalize to unseen conditions beyond its training distribution. To rigorously evaluate this, we tested our pre-trained MM-STT model on six distinct geographical regions, withheld from the training and validation sets, that exhibit a wide spectrum of geophysical behaviors. These regions were specifically selected to represent three challenging categories: (1) continuous, long-term subsidence, (2) stable and periodic variations, and (3) abrupt, large-magnitude co-seismic displacements.

We assess the model's generalization performance from three complementary perspectives, providing a comprehensive view of its capabilities and limitations:
\begin{itemize}
    \item \textbf{Quantitative Performance:} We first evaluate the aggregate forecast accuracy using standard metrics (RMSE, MAE, R²) to establish a baseline for the model's performance across these new and diverse regimes.   
    \item \textbf{Macro-Level Spatial Fidelity:} We then qualitatively analyze the predicted displacement maps at the t+10 horizon. This serves to verify whether the model can correctly capture the large-scale spatial structure and shape of different deformation patterns.
    \item \textbf{Micro-Level Temporal Dynamics:} Finally, we zoom into individual measurement points to scrutinize the model's ability to replicate the precise temporal evolution of various signals, including sharp turning points and post-event stability.
\end{itemize}

\paragraph{Study Sites and Selection Criteria}
To ensure a comprehensive and challenging evaluation of our model's generalization capabilities, we selected six study sites from the European Ground Motion Service (EGMS) dataset that were explicitly excluded from the model's training and validation sets. The selection was guided by the objective to test the model against three distinct categories of ground motion:
\begin{itemize}
    \item \textbf{Continuous and Periodic Deformation:} Regions exhibiting steady, long-term subsidence or clear seasonal/periodic oscillations.
    \item \textbf{Abrupt Co-seismic Displacements:} Regions that experienced sudden, large-magnitude surface displacement due to significant seismic events.
    \item \textbf{Stable Areas:} Regions with minimal deformation, used to test the model's stability in low-signal conditions.
\end{itemize}
Each site is identified using the standard EGMS tiling grid nomenclature, \textbf{EXXNYY}, where: \textbf{XX} is the Easting coordinate (in 100\,km units) and \textbf{YY} is the Northing coordinate (in 100\,km units) of the south-west corner of the tile's lower-left pixel.
The six sites used in our generalization experiments are:
\begin{itemize}
    \item \textbf{E39N30 \& E44N23:} prime examples of \textit{continuous subsidence}.
    \item \textbf{E32N34 \& E32N35:} regions with \textit{periodic variations}, ideal for phase and amplitude forecasting.
    \item \textbf{E48N24 \& E58N17:} areas exhibiting \textit{abrupt co-seismic displacement} (Mw 7.0 Samos and Mw 6.4 Croatia, respectively).
\end{itemize}

\subsubsection{Quantitative Performance on Diverse Regimes}
\label{sssec:quant_generalization}
The quantitative generalization performance of our MM-STT model is summarized in Table~\ref{tab:generalization_results}. The results are presented for two distinct categories of geological phenomena: standard continuous/periodic deformation and challenging abrupt co-seismic events.

In regions characterized by continuous subsidence and periodic variations, our model demonstrates exceptional generalization capabilities. It achieves a remarkable average R² of \textbf{0.9981}, indicating a near-perfect replication of the ground truth dynamics. This performance, on par with our primary test set, confirms that the model has learned the underlying physical principles governing these movements, rather than merely memorizing spatial patterns from a specific training area. This underscores its reliability for long-term monitoring tasks in unseen regions.

Forecasting abrupt, large-magnitude seismic events presents a fundamentally different and more challenging problem. As anticipated, the absolute error metrics (e.g., an average RMSE of 0.8254) are higher in these scenarios, as our model is not designed to predict the stochastic timing of earthquakes. However, a crucial insight emerges from the R² values, which remain outstandingly high (average of \textbf{0.9953}). This indicates that while the model cannot foresee the event itself, it accurately captures the \textit{consequence} of the event. Once the co-seismic displacement is registered in the input sequence, the model correctly models the new spatial deformation field with high fidelity. This intelligent, adaptive behavior effectively defines the model's operational boundary: it excels at pattern-based forecasting of post-event states but is not an earthquake predictor. This finding also points to a clear avenue for future work: integrating real-time seismic precursors to potentially forecast such ``black swan'' events.

\begin{table}[ht]
\centering
\begin{tabular}{l l c c c}
\toprule
\textbf{Deformation Type} & \textbf{Region ID} & \textbf{RMSE} & \textbf{MAE} & \textbf{R²} \\
\midrule
\hline
\multicolumn{5}{l}{\textit{\textbf{Continuous and Periodic Deformation Regimes}}} \\
\hline
Continuous Subsidence & E39N30 & 0.1827 & 0.1434 & 0.9973 \\
Continuous Subsidence & E44N23 & 0.0845 & 0.0676 & 0.9993 \\
Periodic Variation & E32N34 & 0.0821 & 0.0589 & 0.9996 \\
Periodic Variation & E32N35 & 0.1531 & 0.1130 & 0.9963\\
\cmidrule{2-5}
\multicolumn{1}{r}{\textit{Average Performance}} & & \textbf{0.1006} & \textbf{0.0957} & \textbf{0.9981} \\
\hline
\hline
\multicolumn{5}{l}{\textit{\textbf{Co-seismic / Abrupt Displacement Regimes}}} \\
\hline
Co-seismic Step (Mw 7.0) & E48N24 & 0.4236 & 0.2555 & 0.9941 \\
Co-seismic Step (Mw 6.4) & E58N17 & 1.2271 & 0.8080 & 0.9964 \\
\cmidrule{2-5}
\multicolumn{1}{r}{\textit{Average Performance}} & & \textbf{0.8254} & \textbf{0.5318} & \textbf{0.9953} \\
\bottomrule
\end{tabular}
\caption{Generalization Performance of the MM-STT Model Across Diverse Geological Regions.}
\label{tab:generalization_results}
\end{table}

\subsubsection{Qualitative Visualization of Generalization}
\label{sssec:qual_maps}
\begin{figure*}[htbp] 
    \centering 
    \begin{subfigure}[t]{0.32\textwidth}
        \centering
        \includegraphics[width=\linewidth]{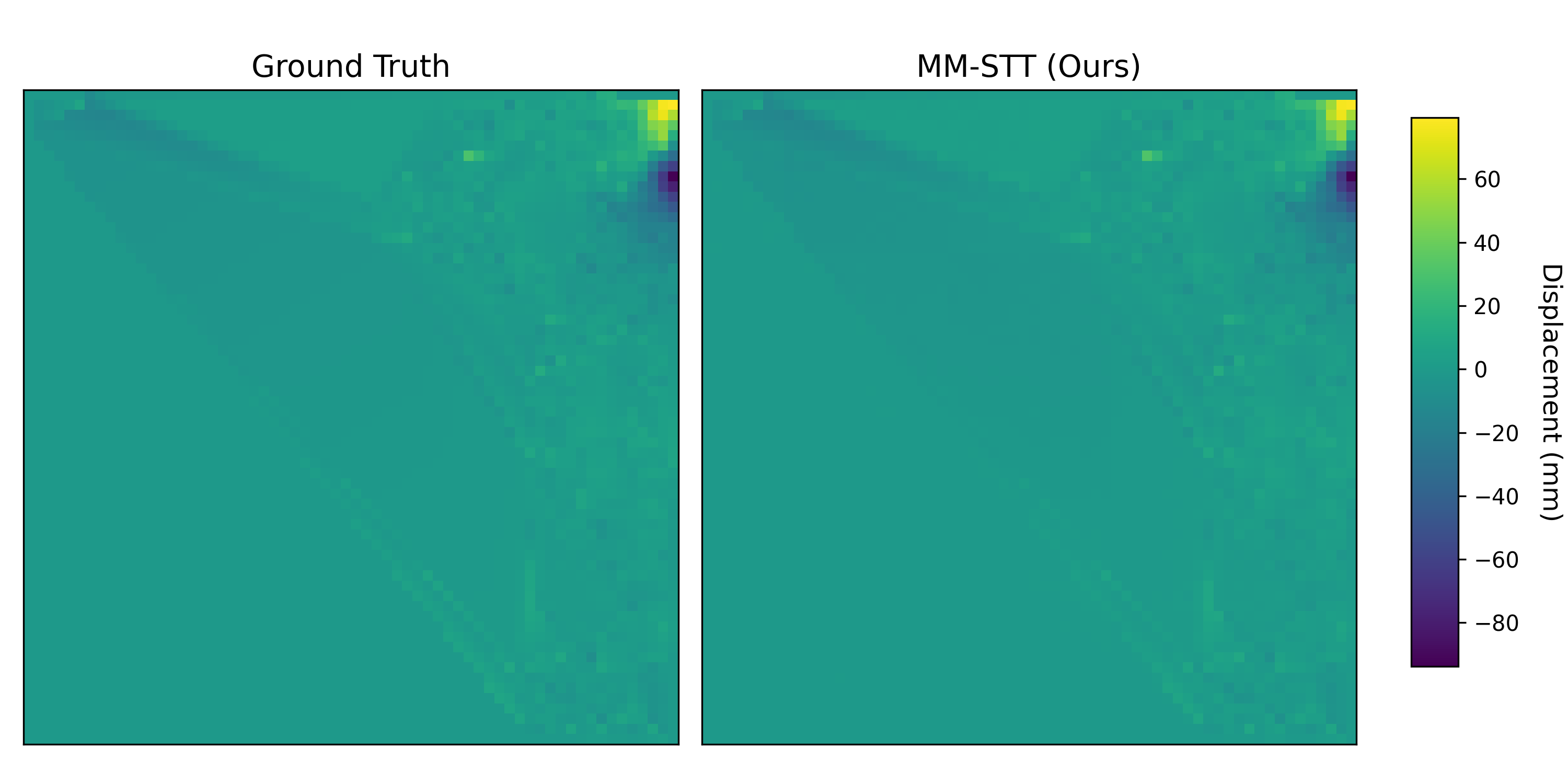}
        \caption{Co-seismic Step (Mw 7.0)}
        \label{fig:map_e48n24}
    \end{subfigure}
    \hfill 
    \begin{subfigure}[t]{0.32\textwidth}
        \centering
        \includegraphics[width=\linewidth]{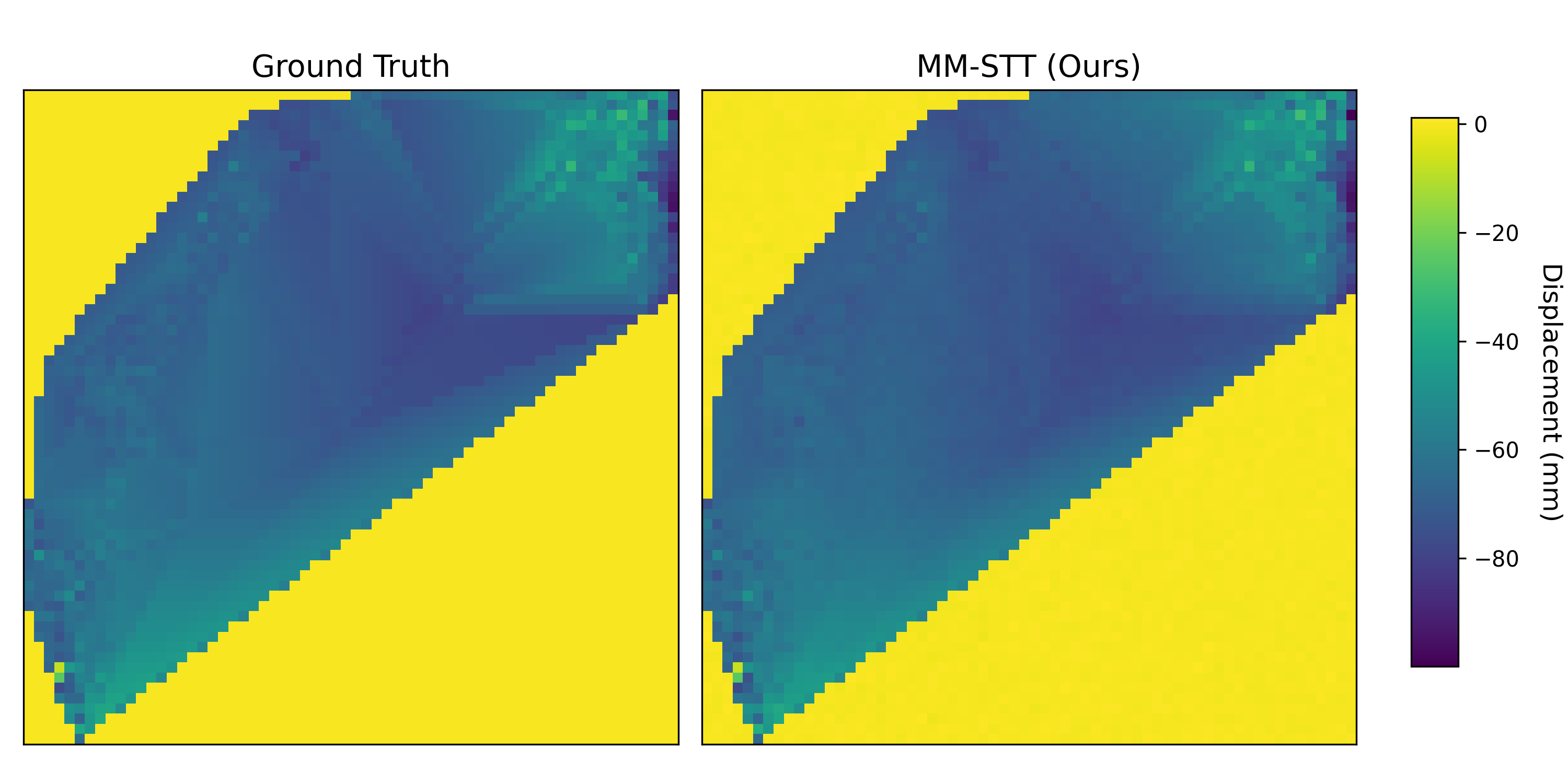}
        \caption{Co-seismic Step (Mw 6.4)}
        \label{fig:map_e58n17}
    \end{subfigure}
    \hfill
    \begin{subfigure}[t]{0.32\textwidth}
        \centering
        \includegraphics[width=\linewidth]{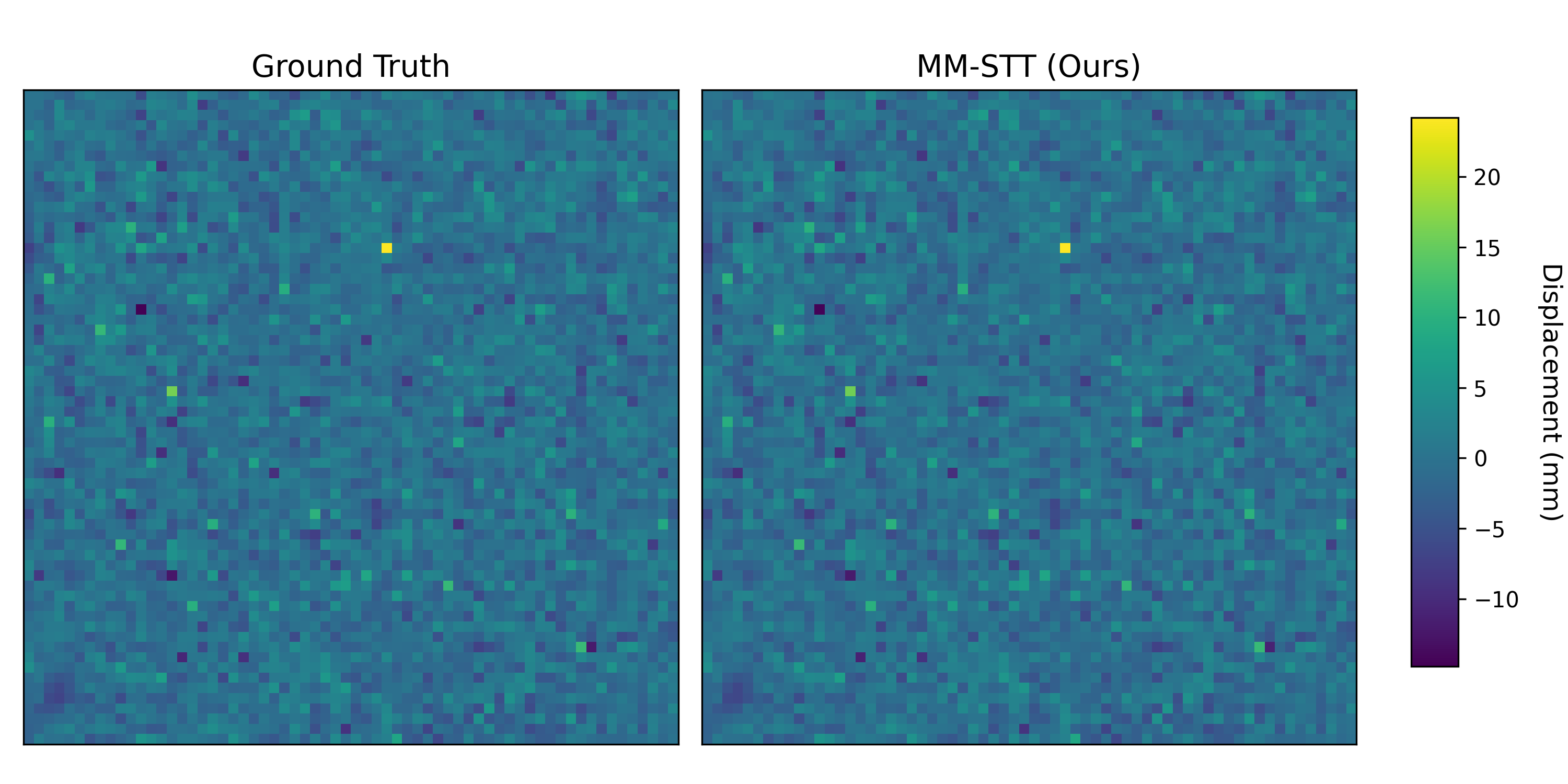}
        \caption{Continuous Subsidence}
        \label{fig:map_e39n30}
    \end{subfigure}
    \begin{subfigure}[t]{0.32\textwidth}
        \centering
        \includegraphics[width=\linewidth]{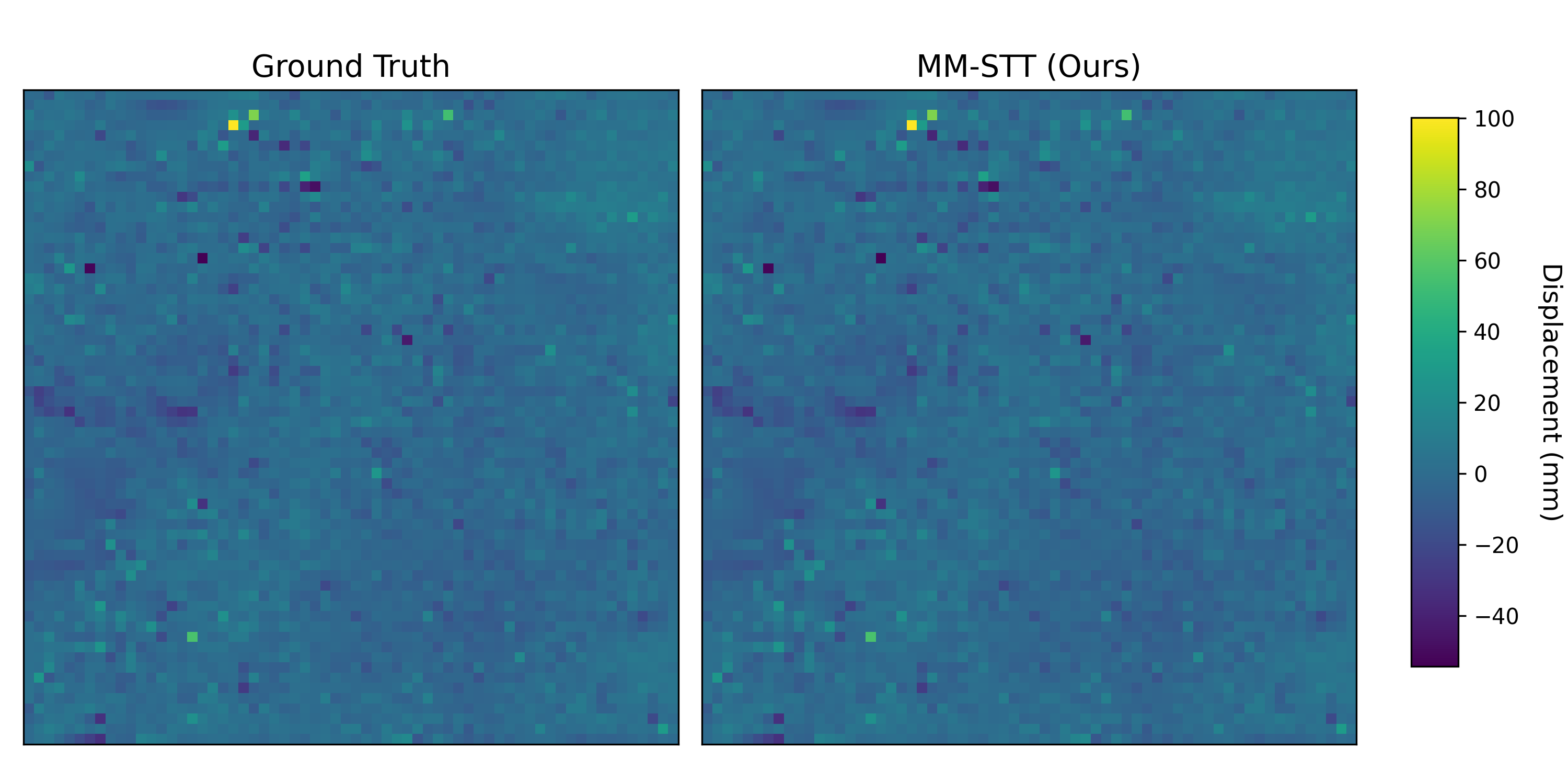}
        \caption{Continuous Subsidence}
        \label{fig:map_e44n23}
    \end{subfigure}
    \hfill
    \begin{subfigure}[t]{0.32\textwidth}
        \centering
        \includegraphics[width=\linewidth]{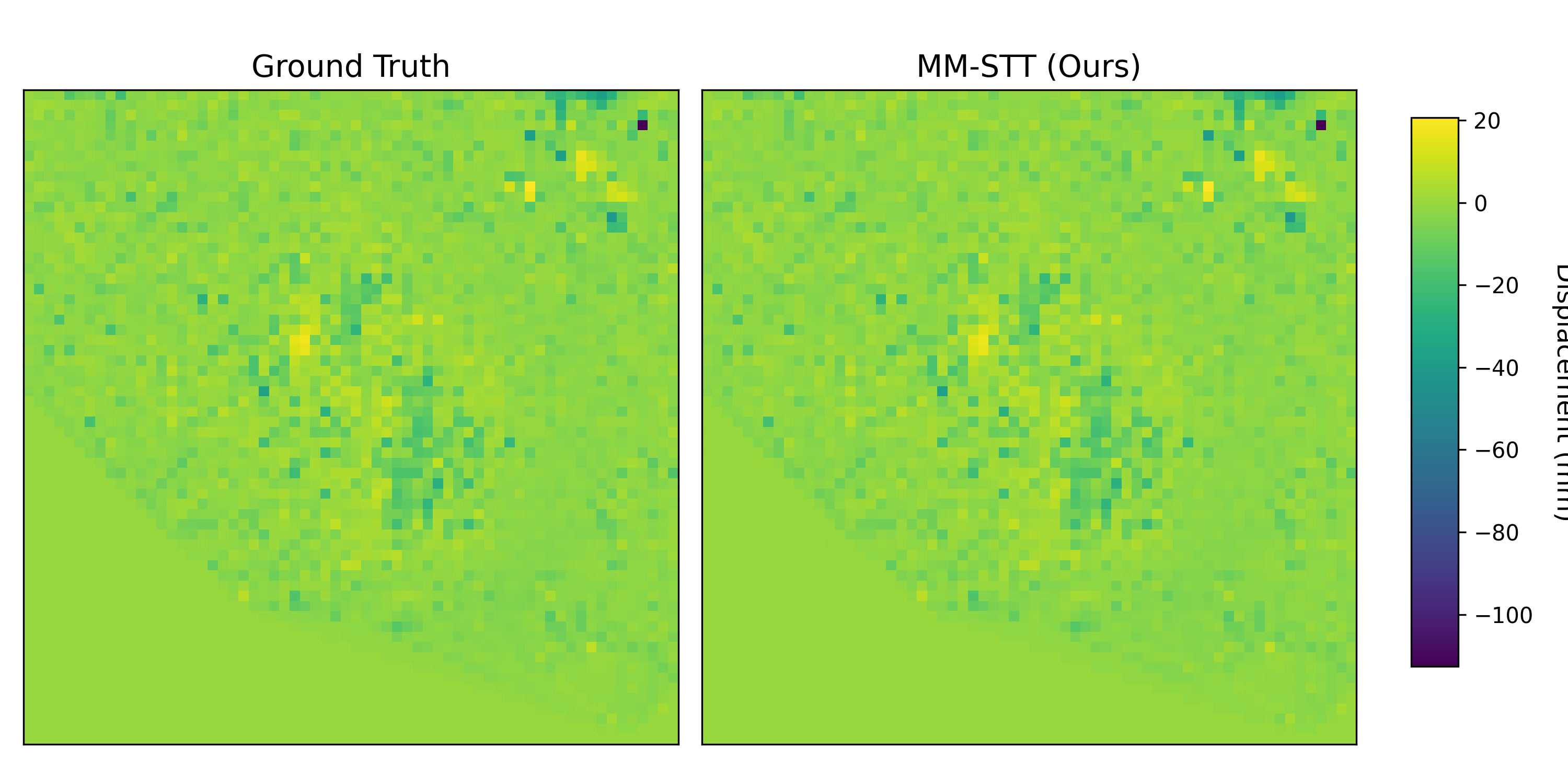}
        \caption{Periodic Variation}
        \label{fig:map_e32n34}
    \end{subfigure}
    \hfill
    \begin{subfigure}[t]{0.32\textwidth}
        \centering
        \includegraphics[width=\linewidth]{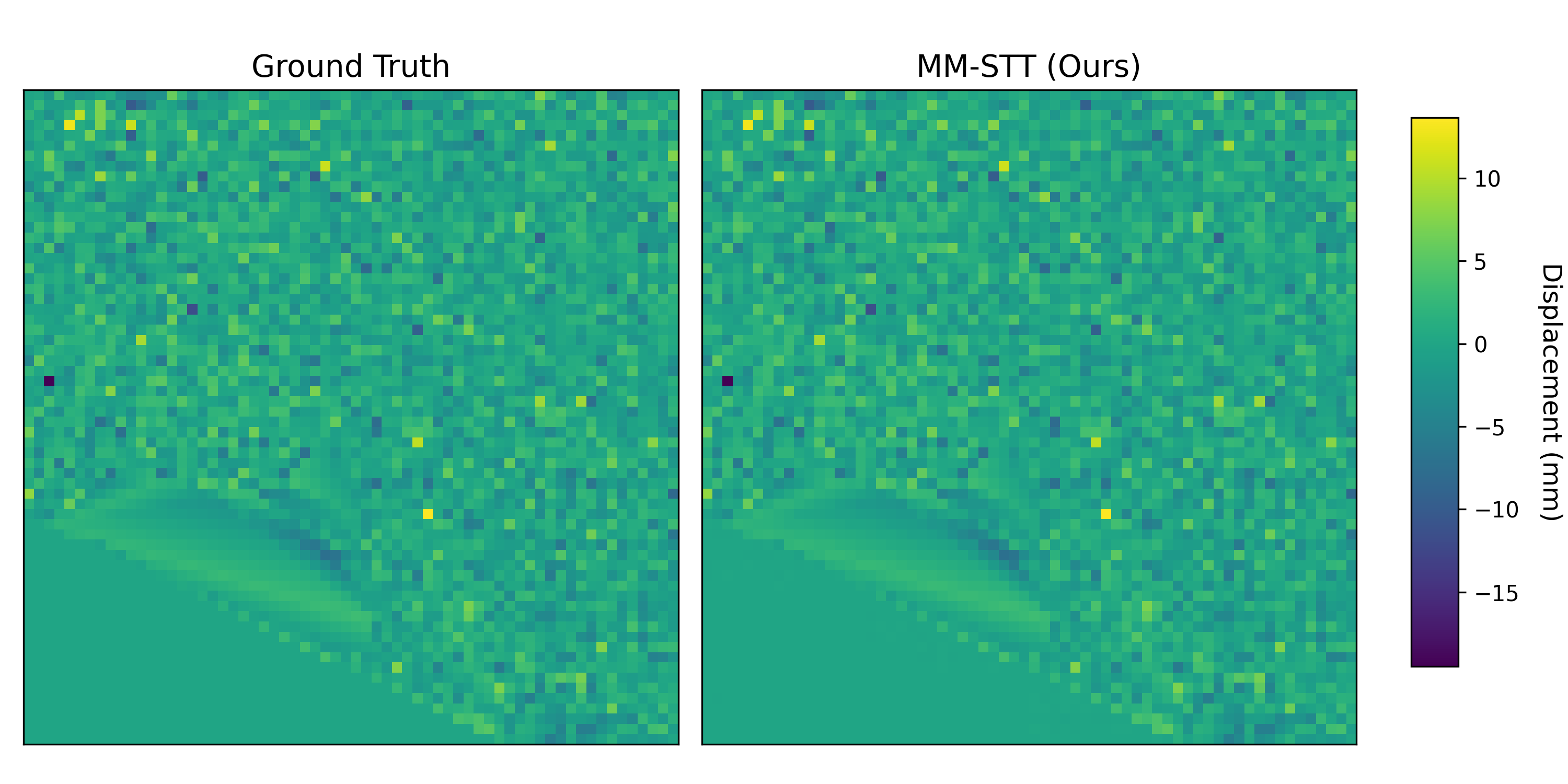}
        \caption{Periodic Variation}
        \label{fig:map_e32n35}
    \end{subfigure}
    \caption{Qualitative comparison of predicted displacement maps at the t+10 forecast horizon across six diverse regions. Each sub-figure compares the Ground Truth (left) against our MM-STT prediction (right).}
    \label{fig:all_displacement_maps} 
\end{figure*}

Beyond quantitative metrics, a qualitative assessment of the predicted displacement maps is crucial for understanding the model's practical capabilities. Figure~\ref{fig:all_displacement_maps} presents a side-by-side comparison of the ground truth and our MM-STT's predictions at the challenging t+10 forecast horizon across six diverse generalization regions.

The visual evidence is striking and strongly corroborates our quantitative findings. Across all scenarios, our model demonstrates remarkable visual fidelity, accurately capturing the key spatial characteristics of the deformation fields. It precisely delineates the sharp boundaries of \textbf{abrupt co-seismic events} (Figs.~\ref{fig:map_e48n24}, \ref{fig:map_e58n17}), reproduces both large-scale features and subtle \textbf{localized ``hotspots''} in continuous and periodic regimes (Figs.~\ref{fig:map_e39n30}, \ref{fig:map_e44n23}, \ref{fig:map_e32n34}, \ref{fig:map_e32n35}), and maintains high \textbf{structural and textural fidelity}, avoiding the overly smooth predictions characteristic of traditional models.

This superior performance is directly attributed to our model's architecture. The \textbf{global self-attention mechanism} allows it to process the entire spatial field as a cohesive whole, modeling large-scale dependencies, while the fusion of \textbf{multi-modal features} provides essential physical context. In summary, the qualitative results confirm that our MM-STT generates spatially coherent and physically plausible forecasts, making it a powerful and reliable tool for real-world geophysical monitoring.

\subsubsection{Micro-Level Temporal Dynamics Analysis}
\label{sssec:micro_temporal}
\begin{figure*}[htbp]
    \centering
    \begin{subfigure}[t]{0.24\textwidth}
        \centering
        \includegraphics[width=\linewidth]{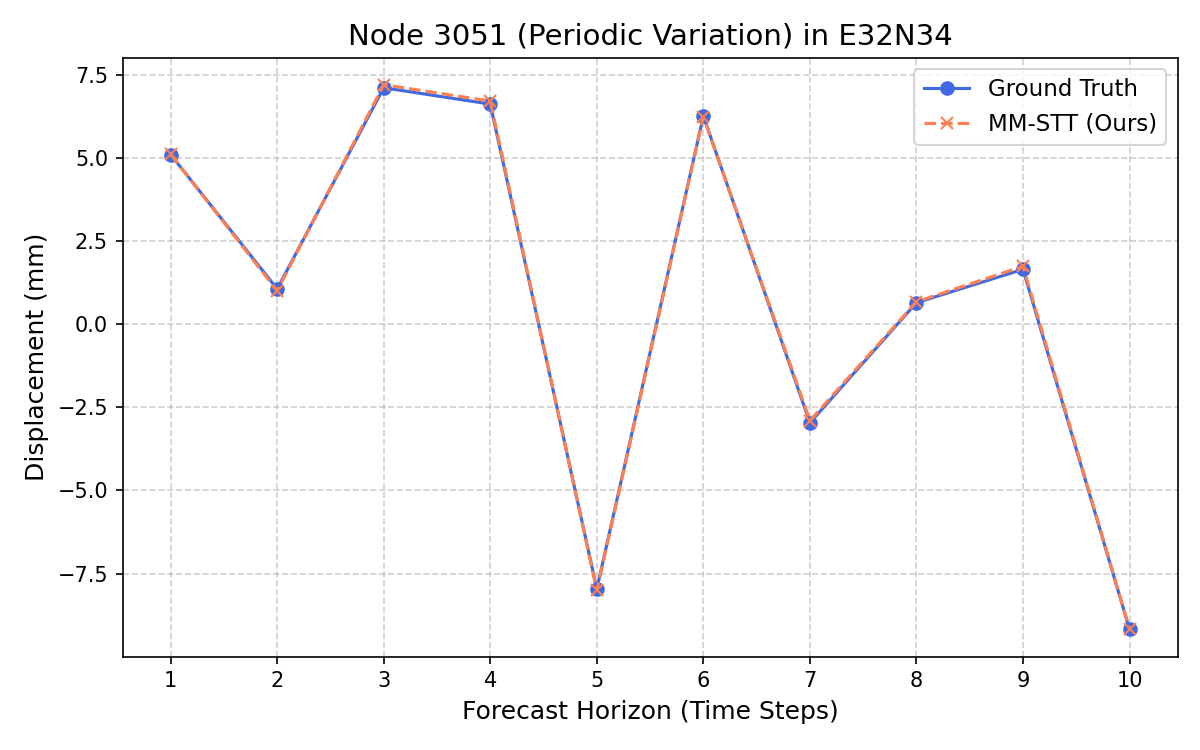}
        \caption{Periodic Variation}
        \label{fig:ts_periodic}
    \end{subfigure}
    \hfill
    \begin{subfigure}[t]{0.24\textwidth}
        \centering
        \includegraphics[width=\linewidth]{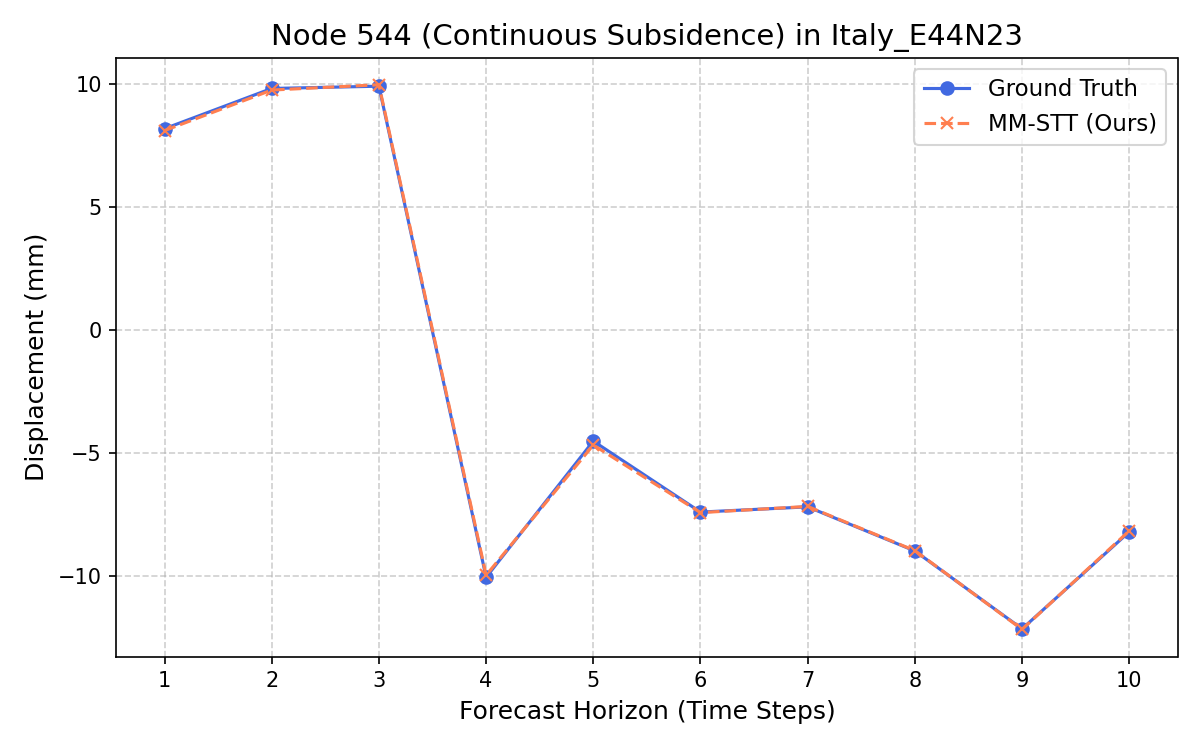}
        \caption{Continuous Subsidence (Non-linear)}
        \label{fig:ts_continuous}
    \end{subfigure}
    \hfill
    \begin{subfigure}[t]{0.24\textwidth}
        \centering
        \includegraphics[width=\linewidth]{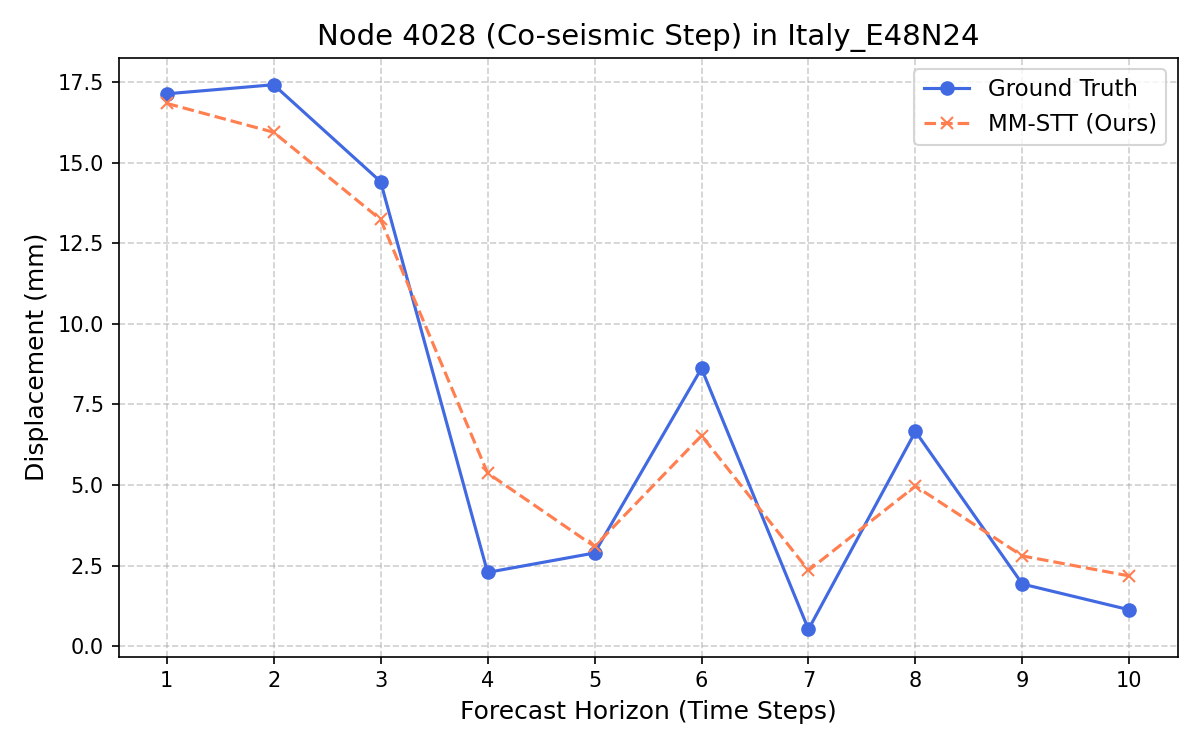}
        \caption{Co-seismic Step}
        \label{fig:ts_coseismic}
    \end{subfigure}
    \hfill
    \begin{subfigure}[t]{0.24\textwidth}
        \centering
        \includegraphics[width=\linewidth]{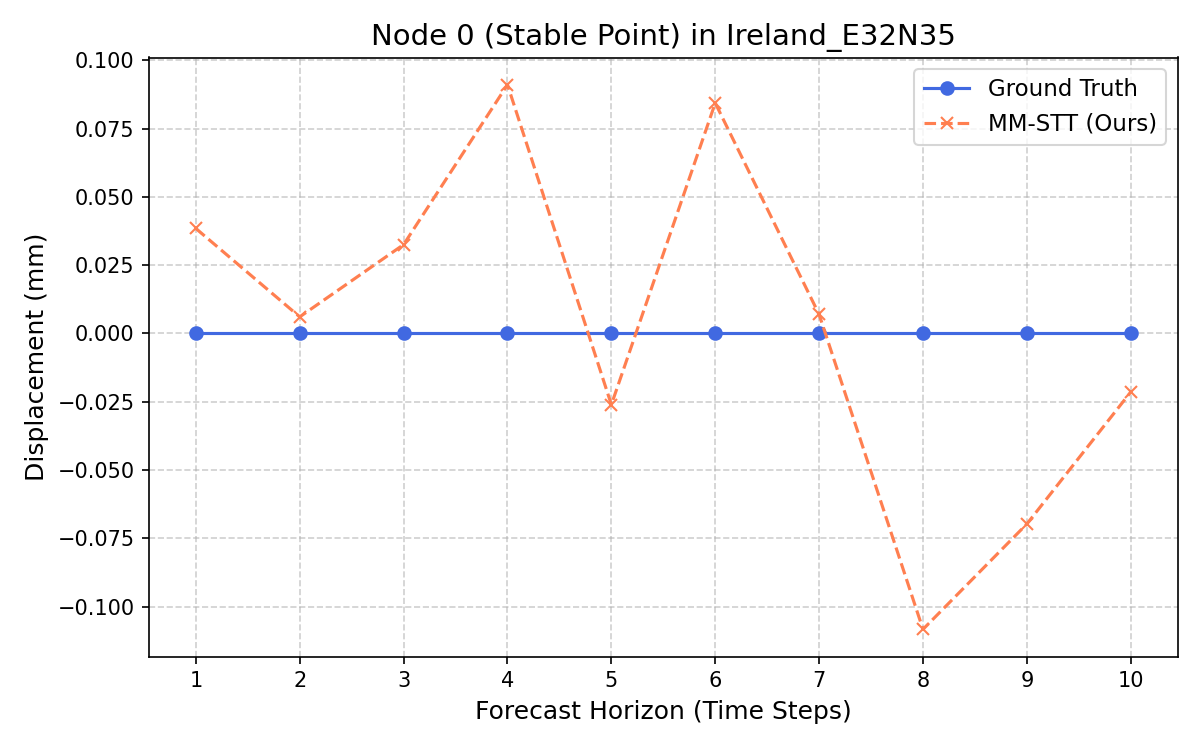}
        \caption{Stable (Near-Zero) Point}
        \label{fig:ts_stable}
    \end{subfigure}
    \caption{Temporal dynamics for four representative nodes from different regions, showing 10-step-ahead forecasts (MM-STT vs. Ground Truth).}
    \label{fig:appendix_timeseries}
\end{figure*}

To provide a deeper insight beyond aggregate metrics, we scrutinize the model's predictive behavior at the level of individual time series. Figure~\ref{fig:appendix_timeseries} showcases the 10-step-ahead forecast for four representative nodes, each selected to highlight the model's performance on a distinct geophysical pattern.

The results are highly illustrative. For standard deformation patterns like periodic variations (Fig.~\ref{fig:appendix_timeseries}a) and non-linear subsidence (Fig.~\ref{fig:appendix_timeseries}b), the model achieves near-perfect accuracy, meticulously tracking the ground truth's trend, phase, and sharp turning points. In more challenging scenarios, the model exhibits nuanced, adaptive behavior: it correctly captures the post-event trend after a co-seismic step (Fig.~\ref{fig:appendix_timeseries}c) and maintains a stable, drift-free forecast for near-zero points (Fig.~\ref{fig:appendix_timeseries}d). The minor oscillations observed in the stable case suggest a high sensitivity to input noise, defining a boundary condition for low-signal environments. Collectively, this micro-level analysis confirms that the strong quantitative performance of our MM-STT is grounded in its robust ability to accurately trace diverse and complex temporal evolutions.

\section{Conclusion}
\label{sec:conclusion}

In this paper, we address the challenging task of predicting high-resolution land subsidence. We demonstrated that a powerful model architecture alone is insufficient and that integrating domain-specific knowledge is paramount. Our proposed multi-modal spatial-temporal transformer (MM-STT) model, which fuses dynamic, static, and temporal data, proved to be highly effective. It successfully captured complex, nonlinear spatiotemporal patterns, significantly outperforming conventional deep learning models. This work underscores the potential of combining rich, multi-modal geoscience data with advanced deep learning architectures for more accurate and reliable geophysical forecasting. Future work will focus on exploring more sophisticated attention mechanisms and applying the framework to larger-scale datasets.

\bibliographystyle{IEEEtran}
\bibliography{ref}

\vfill

\end{document}